\documentclass{bmvc2k}

\title{A Unified Deep Framework for Joint 3D Pose Estimation and Action Recognition from a Single RGB Camera}

\addauthor{Huy Hieu PHAM$^{1,2,}$}{huy-hieu.pham@cerema.fr}{5}
\addauthor{Houssam Salmane}{houssam.salmane@cerema.fr}{1}
\addauthor{Louahdi~Khoudour}{louahdi.khoudour@cerema.fr}{1}
\addauthor{Alain~Crouzil}{alain.crouzil@irit.fr}{2}
\addauthor{Pablo~Zegers}{pablozegers@gmail.com}{3}
\addauthor{Sergio~A.~Velastin}{sergio.velastin@ieee.org}{4}

\addinstitution{
 Cerema Research Center, Toulouse, France
}
\addinstitution{
 Informatics Research Institute of Toulouse (IRIT), Toulouse, France
}

\addinstitution{
 Aparnix, Chile
}

\addinstitution{
 Cortexica Vision Systems Ltd., London, UK
}

\addinstitution{
Vingroup Big Data Institute (VinBDI), Hanoi, Vietnam
}

\runninghead{A Preprint}{Hieu Pham \textit{et al.} 2019}

\def\etal{\emph{et al}\bmvaOneDot}
\usepackage{amssymb}
\usepackage{multirow}
\usepackage[table]{xcolor}
\usepackage{array}
\usepackage{float}
\usepackage{mathtools}

\begin{document}
\maketitle

\begin{abstract}
We present a deep learning-based multitask framework for joint 3D human pose estimation and action recognition from RGB video sequences. Our approach proceeds along two stages. In the first, we run a real-time 2D pose detector to determine the precise pixel location of important keypoints of the body. A two-stream neural network is then designed and trained to map detected 2D keypoints into 3D poses. In the second, we deploy the Efficient Neural Architecture Search (ENAS) algorithm to find an optimal network architecture that is used for modeling the spatio-temporal evolution of the estimated 3D poses via an image-based intermediate representation and performing action recognition. Experiments on Human3.6M, MSR Action3D and SBU Kinect Interaction datasets verify the effectiveness of the proposed method on the targeted tasks. Moreover, we show that our method requires a low computational budget for training and inference.
\end{abstract} 

\section{Introduction \\[-0.2cm]}\label{sec:intro} 
Human action recognition from videos has been researched for decades, since this topic plays a key role in various areas such as intelligent surveillance, human-robot interaction, robot vision and so on. Although significant progress has been achieved in the past few years, building an accurate, fast and efficient system for the recognition of actions in unseen videos is still a challenging task due to a number of obstacles, \textit{e.g.} changes in camera viewpoint, occlusions, background, speed of motion, etc. Traditional approaches on video-based action recognition \cite{weinland2011survey} have focused on extracting hand-crafted local features and building motion descriptors from RGB sequences. Many spatio-temporal representations of human motion have been proposed and widely exploited with success such as SIFT \cite{lowe2004distinctive}, HOF \cite{laptev2008learning} or Cuboids \cite{dollar2005behavior}. However, one of the major limitations of these approaches is the lack of 3D structure from the scene and recognizing human actions based only on RGB information is not enough to overcome the current challenges of the field. 

The rapid development of depth-sensing time-of-flight camera technology has helped in dealing with this problem, which is considered complex for traditional cameras. Low-cost and easy-to-use depth cameras are able to provide detailed 3D structural information of human motion. In particular, most of the current depth cameras have integrated real-time skeleton estimation and tracking frameworks \cite{ye2014real}, facilitating the collection of skeletal data. This is a high-level representation of the human body, which is suitable for the problem of motion analysis.  Hence, exploiting skeletal data for 3D action recognition opens up opportunities for addressing the limitations of RGB-based solutions and many skeleton-based action recognition approaches have been proposed \cite{Wang2012MiningAE,xia2012view,chaudhry2013bio,Vemulapalli2014HumanAR,ding2016profile}. However, depth sensors have some significant drawbacks with respect to 3D pose estimation. For instance, they are only able to operate up to a limited distance and within a limited field of view. Moreover, a major drawback of depth cameras is the inability to work in bright light, especially sunlight \cite{zhang2012microsoft}.
\begin{figure}[h]
\begin{center}
 \includegraphics[width=9.5cm,height=3.5cm]{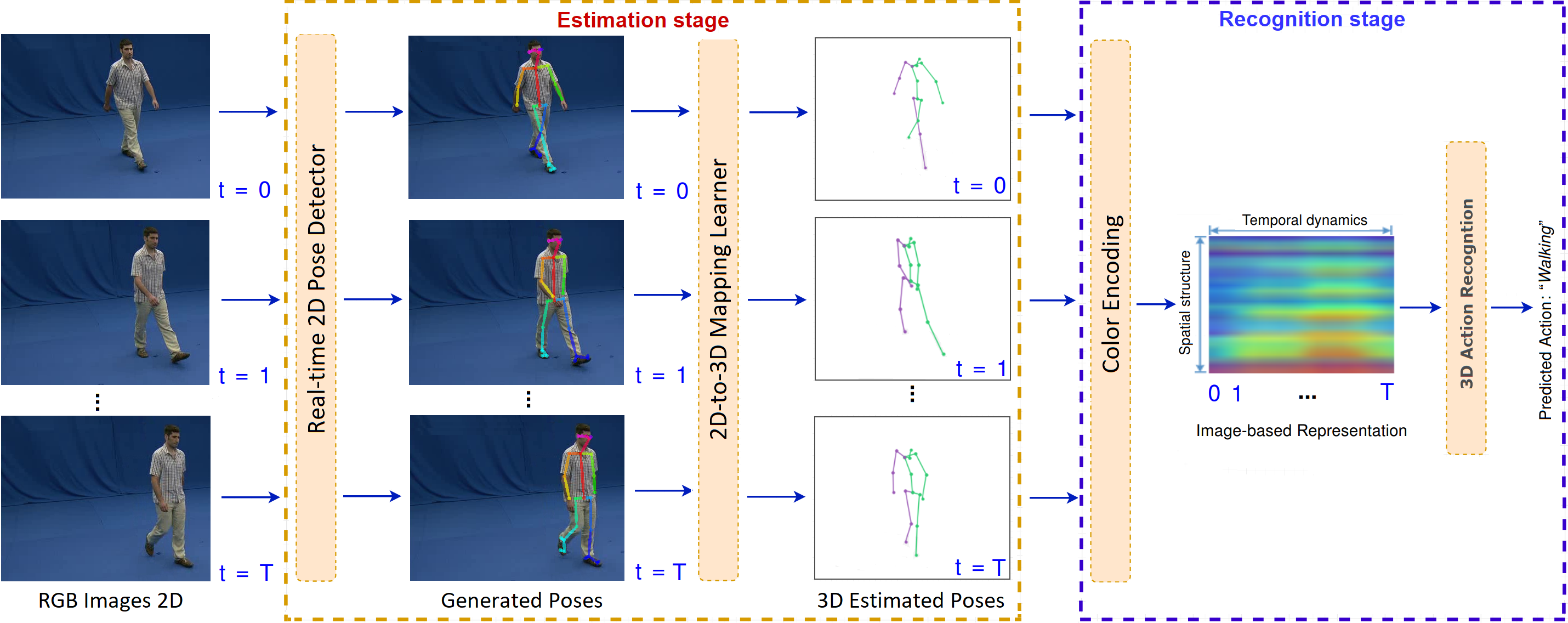}
 \vspace*{-0.4cm}
\end{center}
\caption{Overview of the proposed method. In the estimation stage, we first run OpenPose \cite{cao2017realtime} -- a real-time, state-of-the-art multi-person 2D pose detector to generate 2D human body keypoints. A deep neural network is then trained to produce 3D poses from the 2D detections. In the recognition stage, the 3D estimated poses are encoded into a compact image-based representation and finally fed into a deep convolutional network for supervised classification task, which is automatically searched by the ENAS algorithm \cite{pmlr-v80-pham18a}.}
\label{fig:1}
\end{figure}
Our focus in this paper is therefore to propose a 3D skeleton-based action recognition approach without depth sensors. Specifically, we are interested in building a unified deep framework for both 3D pose estimation and action recognition from RGB video sequences. As shown in Figure \ref{fig:1}, our approach consists of two stages. In the first, \textit{estimation stage}, the system recovers the 3D human poses from the input RGB video. In the second, \textit{recognition stage}, an action recognition approach is developed and stacked on top of the 3D pose estimator in a unified framework, where the estimated 3D poses are used as inputs to learn the spatio-temporal motion features and predict action labels. 

There are four hypotheses that motivate us to build a deep learning framework for human action recognition from 3D poses. First, actions can be correctly represented through the 3D pose movements \cite{johansson1975visual,gu2010action}. Second, the 3D human pose has a high-level of abstraction with much less complexity compared to RGB and depth streams. This makes the training and inference processes much simpler and faster. Third, depth cameras are able to provide highly accurate skeletal data for 3D action recognition. However, they are expensive and not always available (\textit{e.g.} for outdoor scenes). A fast and accurate approach of 3D pose estimation from only RGB input is highly desirable. Fourth, state-of-the-art 2D pose detectors \cite{cao2017realtime,10.1007/978-3-319-46484-8_29} are able to provide 2D poses with a high degree of accuracy in real-time. Meanwhile, deep networks have proved their capacity to learn complex functions from high-dimensional data. Hence, a simple network model can also learn a \textit{mapping} to convert 2D poses into 3D. 

The effectiveness of the proposed method is evaluated on public benchmark datasets (Human3.6M \cite{6682899}, MSR Action3D \cite{li2010action}, and SBU \cite{yun2012two}). Far beyond our expectations, the experimental results demonstrate state-of-the-art performances on the targeted tasks (Section \ref{sect:4.3}) and support our hypotheses above. Furthermore, we show that this approach has a low computational cost (Section \ref{sect:4.4}). Overall, our main contributions are as follows: \\[0.15cm]
\hspace*{0.5cm} $\bullet$ First, we present a two-stream, lightweight neural network to recover 3D human poses from RGB images provided by a monocular camera. Our proposed method achieves state-of-the-art result on 3D human pose estimation task and benefits action recognition.\\[0.15cm]
\hspace*{0.5cm} $\bullet$ Second, we propose to put an action recognition approach on top of the 3D pose estimator to form a unified framework for 3D pose-based action recognition. It takes the 3D estimated poses as inputs, encodes them into a compact image-based representation and finally feeds to a deep convolutional network, which is designed automatically by using a neural architecture search algorithm. Surprisingly, our experiments show that we reached state-of-the-art results on this task, even when compared with methods using depth cameras.\\[0.15cm]
The rest of this paper is organized as follows. We present a review of the related work in Section \ref{sect:2}. The proposed method is explained in Section \ref{sect:3}. Experiments are provided in Section \ref{sect:4} and Section \ref{sect:5} concludes the paper.\\[-0.75cm]
\section{Related work \\[-0.2cm]} \label{sect:2}
This section reviews two main topics that are directly related to ours, \textit{i.e.} 3D pose estimation from RGB images and 3D pose-based action recognition. Due to the limited size of a conference paper, an extensive literature review is beyond the scope of this section. Instead, the interested reader is referred to the surveys of Sarafianos \etal \cite{Sarafianos20163DHP} for recent advances in 3D human pose estimation and Presti \etal \cite{presti20163d} for 3D skeleton-based action recognition.\\[-0.8cm]
\subsection{3D human pose estimation \\[-0.1cm]}
The problem of 3D human pose estimation has been intensively studied in the recent years. Almost all early approaches for this task were based on feature engineering \cite{4020735,10.1007/978-3-642-33765-9_41,6682899}, while the current state-of-the-art methods are based on deep neural networks \cite{li20143d,tekin2016direct,pavlakos2017coarse,pavllo20183d,VNect_SIGGRAPH2017,Katircioglu2018}. Many of them are regression-based approaches that directly predict 3D poses from RGB images via 2D/3D heatmaps. For instance, Li \etal \cite{li20143d} designed a deep convolutional network for human detection and pose regression. The regression network learns to predict 3D poses from single images using the output of a body
part detection network. Tekin \etal \cite{tekin2016direct} proposed to use a deep network to learn a regression mapping that directly estimates the 3D pose in a given frame of a sequence from a spatio-temporal volume centered on it. Pavlakos \etal \cite{pavlakos2017coarse} used multiple fully convolutional networks to construct a volumetric stacked hourglass architecture, which is able to recover 3D poses from RGB images. Pavllo \etal \cite{pavllo20183d} exploited a temporal dilated convolutional network \cite{Yu2016MultiScaleCA} for estimating 3D poses. However, this approach led to a significant increase in the number of parameters as well as the required memory. Mehta \etal \cite{VNect_SIGGRAPH2017} introduced a real-time approach to predict 3D poses from a single RGB camera. They used ResNets \cite{7780459} to jointly predict 2D and 3D heatmaps as regression tasks. Recently, Katircioglu \etal \cite{Katircioglu2018} introduced a deep regression network for predicting 3D human poses from monocular images via 2D joint location heatmaps. This architecture is in fact an overcomplete autoencoder that learns a high-dimensional latent pose representation and accounts for joint dependencies, in which a Long Short-Term Memory (LSTM) network \cite{Hochreiter1997LongSM} is used to enforce temporal consistency on 3D pose predictions.

To the best of our knowledge, several studies \cite{pavlakos2017coarse,VNect_SIGGRAPH2017,Katircioglu2018} stated that regressing the 3D pose from 2D joint locations is difficult and not too accurate. However, motivated by Martinez \etal \cite{martinez_2017_3dbaseline}, we believe that a simple neural network can learn effectively a \textit{direct 2D-to-3D mapping}. Therefore, this paper aims at proposing a simple, effective and real-time approach for 3D human pose estimation that benefits action recognition. To this end, we design and optimize a two-stream deep neural network that performs 3D pose predictions from the 2D human poses. These 2D poses are generated by a state-of-the-art 2D detector that is able to run in real-time for multiple people. We empirically show that although the proposed approach is computationally inexpensive, it is still able to improve the state-of-the-art.\\[-0.8cm]
\subsection{3D pose-based action recognition \\[-0.1cm]}
Human action recognition from skeletal data or 3D poses is a challenging task. Previous works on this topic can be divided into two main groups of method. The first group \cite{Lv2006RecognitionAS,Wang2012MiningAE,Vemulapalli2014HumanAR} extracts hand-crafted features and uses probabilistic graphical models, \textit{e.g.} Hidden Markov Model (HMM) \cite{Lv2006RecognitionAS} or Conditional Random Field (CRF) \cite{Han2010DiscriminativeHA} to recognize actions. However, almost all of these approaches require a lot of feature engineering. The second group \cite{Liu2016SpatioTemporalLW,Du2015HierarchicalRN,Shahroudy2016NTURA} considers the 3D pose-based action recognition as a time-series problem and proposes to use Recurrent Neural Networks with Long-Short Term Memory units (RNN-LSTMs) \cite{Hochreiter1997LongSM} for modeling the dynamics of the skeletons. Although RNN-LSTMs are able to model the long-term temporal characteristics of motion and have advanced the state-of-the-art, this approach feeds raw 3D poses directly into the network and just considers them as a kind of low-level feature. The large number of input features makes RNNs very complex and may easily lead to overfitting. Moreover, many RNN-LSTMs act merely as classifiers and cannot extract high-level features for recognition tasks \cite{Sainath2015ConvolutionalLS}.

In the literature, 3D human pose estimation and action recognition are closely related. However, both problems are generally considered as two distinct tasks \cite{cheronICCV15}. Although some approaches have been proposed for tackling the problem of jointly predicting 3D poses and recognizing actions in RGB images or video sequences \cite{5540235,7298734,luvizon20182d}, they are data-dependent and require a lot of feature engineering, except the work of Luvizon \etal \cite{luvizon20182d}. Unlike in previous studies, we propose a multitask learning framework for 3D pose-based action recognition by reconstructing 3D skeletons from RGB images and exploiting them for action recognition in a joint way. Experimental results on public and challenging datasets show that our framework is able to solve the two tasks in an effective way. \\[-0.8cm]
\section{Proposed Method\\[-0.2cm]} \label{sect:3}
We explain the proposed method is this section. First, our approach for 3D human pose estimation is presented. We then introduce our solution for 3D pose-based action recognition. \\[-0.8cm]
\subsection{Problem definition \\[-0.1cm]}
Given an RGB video clip of a person who starts to perform an action at time $t = 0$ and ends at $t = T$, the problem studied in this work is to generate a sequence of 3D poses $\mathcal{P} = (\textbf{p}_0, ..., \textbf{p}_T)$, where $\textbf{p}_i \in \mathbb{R}^{3 \times M}$, $i \in \{0, ..., T\}$ at the estimation stage. The generated $\mathcal{P}$ is then used as input for the recognition stage to predict the corresponding action label $\mathcal{A}$ by a supervised learning model. See Figure \ref{fig:1} for an illustration of the problem. \\[-0.8cm]
\subsection{3D human pose estimation \\[-0.1cm]}
Given an input RGB image $\textbf{I} \in \mathbb{R}^{W \times H \times 3}$, we aim to estimate the body joint locations in the 3-dimensional space, noted as $\hat{\textbf{p}}_{3D}$ $\in \mathbb{R}^{3 \times M}$. To this end, we first run the state-of-the-art human 2D pose detector, namely OpenPose \cite{cao2017realtime}, to produce a series of 2D keypoints $\textbf{p}_{2D} \in \mathbb{R}^{2 \times N}$. To recover the 3D joint locations, we try to learn a \textit{direct 2D-to-3D mapping} $f_r$: $\textbf{p}_{2D} \xmapsto{f_r} \hat{\textbf{p}}_{3D}$. This transformation can be implemented by a deep neural network in a supervised manner
\begin{equation}
    \hat{\textbf{p}}_{3D} = f_r(\textbf{p}_{2D}, \theta),
\end{equation}
where $\theta$ is a set of trainable parameters of the function $f_r$. To optimize $f_r$, we minimize the prediction error over a labelled dataset of $\mathcal{C}$ poses by solving the optimization problem
\begin{equation}
   \underset{\theta}{\operatorname{arg\,min}} \hspace{0.2cm}  \frac{1}{\mathcal{C}} \sum_{n=1}^{\mathcal{C}} \mathcal{L} (f_r(\textbf{x}_i), \textbf{y}_i).
\end{equation}
Here $\textbf{x}_i$ and $\textbf{y}_i$ are the input 2D poses and the ground truth 3D poses, respectively; $\mathcal{L}$ denotes a loss function. In our implementation the robust Huber loss \cite{huber1992robust} is used to deal with outliers. \\[-0.7cm]
\subsubsection{Network design}
\begin{figure}[h]
\begin{center}
 \includegraphics[width=10cm,height=4.2cm]{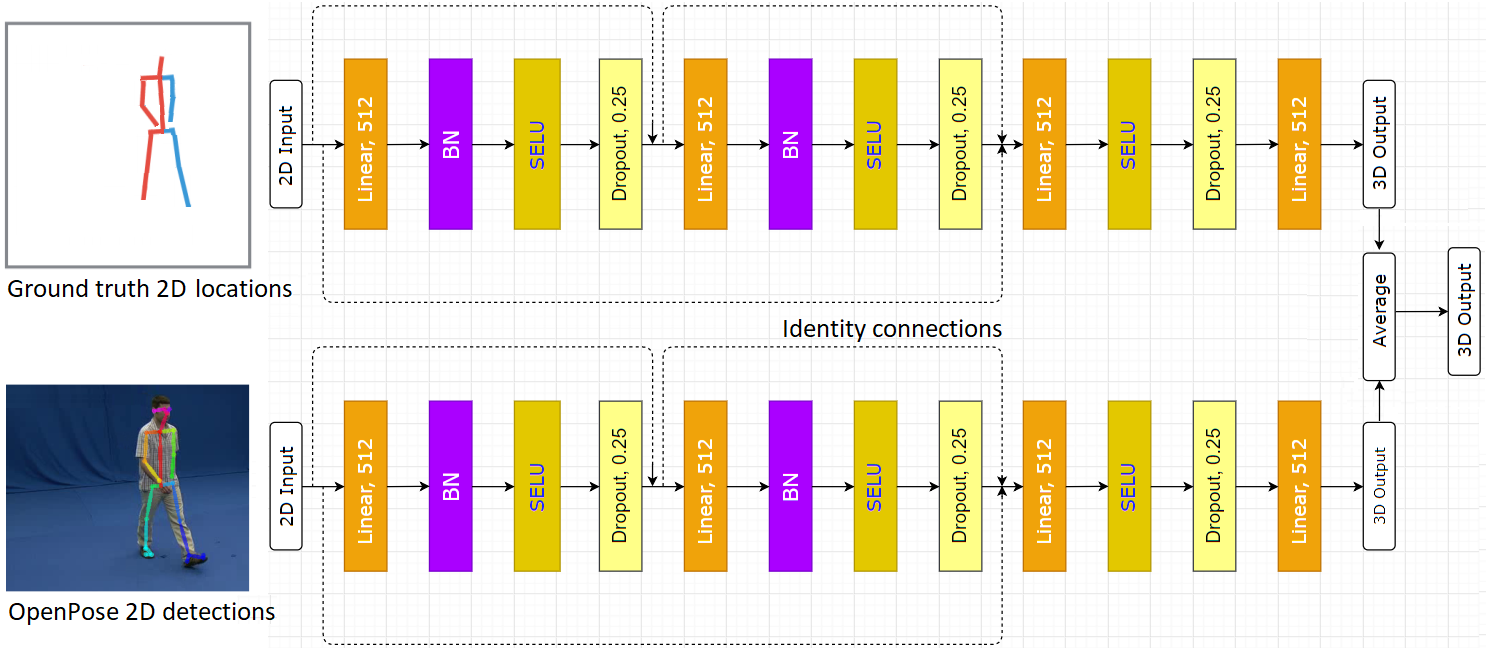} 
 \vspace*{-0.4cm}
\end{center}
   \caption{Diagram of the proposed two-stream network for training our 3D pose estimator.}
\label{fig:2}
\end{figure}

State-of-the-art deep learning architectures such as ResNet \cite{7780459}, Inception-ResNet-v2 \cite{Szegedy2016Inceptionv4IA}, DenseNet \cite{Huang2017DenselyCC}, or NASNet \cite{Zoph2017NeuralAS} have achieved an impressive performance in supervised learning tasks with high dimensional data, \textit{e.g.} 2D or 3D images. However, the use of these architectures \cite{7780459,Szegedy2016Inceptionv4IA,Huang2017DenselyCC,Zoph2017NeuralAS} on low dimensional data like the coordinates of the 2D human joints could lead to overfitting. Therefore, our design is based on a simple and lightweight multilayer network architecture without the convolution operations. In the design process, we exploit some recent improvements in the optimization of the modern deep learning models \cite{7780459,Huang2017DenselyCC}. Concretely, we propose a two-stream network. Each stream comprises linear layers, Batch Normalization (BN) \cite{Ioffe2015BatchNA}, Dropout \cite{Hinton2012ImprovingNN}, SELU \cite{Klambauer2017SelfNormalizingNN} and Identity connections \cite{7780459}. During the training phase, the first stream takes the  ground truth 2D locations as input. The 2D human joints predicted by OpenPose \cite{cao2017realtime} are inputted to the second stream. The outputs of the two streams are then averaged. Figure \ref{fig:2} illustrates our network design. Note that learning with the ground truth 2D locations for both of these streams could lead to a higher level of performance. However, training with the 2D OpenPose detections could improve the generalization ability of the network and makes it more robust during the inference, when only the OpenPose's 2D output is used to deal with action recognition in the wild.
\\[-1cm]
\subsection{3D pose-based action recognition \\[-0.1cm]}
\begin{figure}[ht]
\begin{center}
 \includegraphics[width=10cm,height=1.8cm]{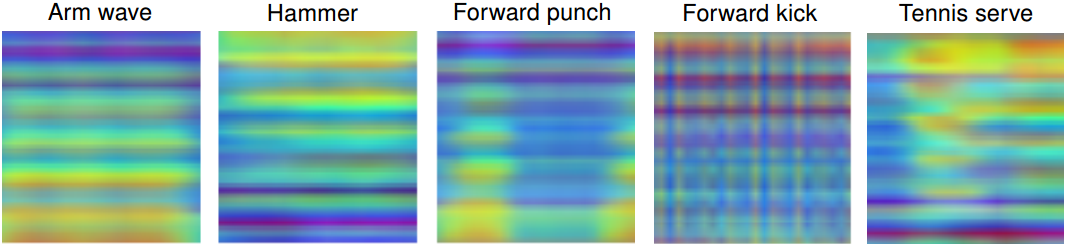} 
 \vspace*{-0.4cm}
\end{center}
   \caption{Immediate image-based representations for the recognition stage.}
\label{fig:3}
\end{figure}

\begin{figure}[ht]
\begin{center}
 \includegraphics[width=13cm,height=2.2cm]{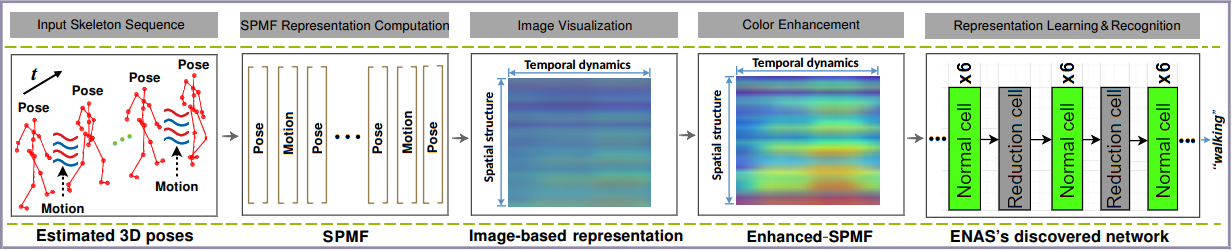}
 \vspace*{-0.8cm}
\end{center}
   \caption{Illustration of the proposed approach for 3D pose-based action recognition.}
\label{fig:4}
\end{figure}

In this section, we explain how to integrate the estimation stage with the recognition stage in a unified framework. Specifically, the proposed recognition approach is stacked on top of the 3D pose estimator. To explore the high-level information of the estimated 3D poses, we encode them into a compact image-based representation. These intermediate representations are then fed to a Deep Convolutional Neural Network (D-CNNs) for learning and classifying actions. This idea has been proven effective in \cite{Pham2018ExploitingDR,Pham2018SkeletalMT,s19081932}. Thus, the spatio-temporal patterns of a 3D pose sequence are transformed into a single color image as a global representation called Enhanced-SPMF \cite{s19081932} via two important elements of a human movement: 3D poses and their motions. Due to the limited space available, detailed description of the Enhanced-SPMF is not included. We refer the interested reader to the work described in \cite{s19081932} for further technical details. Figure \ref{fig:3} visualizes some Enhanced-SPMF representations from samples of the MSR Action3D dataset \cite{li2010action}. 
\\
\hspace*{0.5cm} For learning and classifying the obtained images, we propose to use the Efficient Neural Architecture Search (ENAS) \cite{pmlr-v80-pham18a} -- a recent state-of-the-art technique for automatic design of deep neural networks. The ENAS is in fact an extension of an important advance in deep learning called NAS \cite{Zoph2017NeuralAS}, which is able to automatize the designing process of convolutional architectures on a dataset of interest. This method proposes to search for optimal building blocks (called \textit{cells}, including \textit{normal cells} and \textit{reduction cells}) and the final architecture is then constructed from the best cells achieve. In NAS, an RNN is used. It first samples a candidate architecture called \textit{child model}. This child model is then trained to convergence on the desired task and reports its performance. Next, the RNN uses the performance as a guiding signal to find a better architecture. This process is repeated for many times, making NAS computationally expensive and time-consuming (\textit{e.g.} on CIFAR-10, NAS needs 4 days with 450 GPUs to discover the best architecture). ENAS has been proposed to improve the efficiency of NAS. Its key idea of ENAS \cite{pmlr-v80-pham18a} is the use of sharing parameters among child models, which helps reducing the time of training each child model from scratch to convergence. State-of-the-art performance has been achieved by ENAS on  well known public datasets. We encourage the readers to refer to the original paper \cite{pmlr-v80-pham18a} for more details. Figure \ref{fig:4} illustrates the entire pipeline of our approach for the recognition stage. \\[-0.6cm]
  \\[-1.2cm]
\section{Experiments \\[-0.2cm]} \label{sect:4}
\subsection{Datasets and settings \\[-0.1cm]}
We evaluate the proposed method on three challenging datasets: Human3.6M, MSR Action3D and SBU Kinect Interaction. The Human3.6M is used for evaluating 3D pose estimation. Meanwhile, the other two datasets are used for validating action recognition. The characteristics of each dataset are as follows.\\[0.15cm]
\textbf{Human3.6M} \cite{6682899}: This is a very large-scale dataset containing 3.6 million different 3D articulated poses captured from 11 actors for 17 actions, under 4 different viewpoints. For each subject, the dataset provides 32 body joints, from which only 17 joints are used for training and computing scores. In particular, 2D joint locations and 3D poses ground truth are available for evaluating supervised learning models.\\[0.15cm]
\textbf{MSR Action3D} \cite{li2010action}: This dataset contains 20 actions, performed by 10 subjects. Our experiment was conducted on 557 video sequences of the MSR Action3D, in which the whole dataset is  divided into three subsets: AS1, AS2, and AS3. There are 8 actions classes for each subset. Half of the data is selected for training and the rest is used for testing.\\[0.15cm]
\textbf{SBU Kinect Interaction} \cite{yun2012two}: This dataset contains a total of 300 interactions, performed by 7 participants for 8 actions. This is a challenging dataset due to the fact that it contains pairs of actions that are difficult to distinguish such as \textit{exchanging objects} -- \textit{shaking hands} or \textit{pushing} -- \textit{punching}. We randomly split the whole dataset into 5 folds, in which 4 folds are used for training and the remaining 1 fold is used for testing.\\[-0.7cm]
\subsection{Implementation details \\[-0.1cm]}
The proposed networks were implemented in Python with Keras/TensorFlow backend. The two streams of the 3D pose estimator are trained separately with the same hyperparameters setting, in which we use mini-batches of 128 poses with 0.25 dropout rate. The weights are initialized by the He initialization \cite{He2015DelvingDI}. Adam optimizer \cite{Kingma2014AdamAM} is used with default parameters. The initial learning rate is set to 0.001 and is decreased by a factor of 0.5 after every 50 epochs. The network is trained for 300 epochs from scratch on the Human3.6M dataset \cite{6682899}. For action recognition task, we run OpenPose \cite{cao2017realtime} to generate 2D detections on MSR Action3D \cite{li2010action} and SBU Kinect Interaction \cite{yun2012two}. The pre-trained 3D pose estimator on Human3.6M \cite{6682899} is then used to provide 3D poses. We use standard data pre-processing and augmentation techniques such as randomly cropping and flipping on these two datasets due to their small sizes. To discover optimal recognition networks, we use ENAS \cite{pmlr-v80-pham18a} with the same parameter setting as the original work. Concretely, the shared parameters $\omega$  are trained with Nesterov's accelerated gradient descent \cite{Nesterov} using Cosine learning rate \cite{Loshchilov2016SGDRSG}. The candidate architectures are initialized by He initialization \cite{He2015DelvingDI} and trained by Adam optimizer \cite{Kingma2014AdamAM} with a learning rate of 0.00035. Additionally, each search is run for 200 epochs.\\[-0.75cm]
\subsection{Experimental results and comparison \\[-0.1cm]} \label{sect:4.3}
\subsubsection{Evaluation on 3D pose estimation}
\begin{figure}[ht]
\begin{center}
 \includegraphics[width=12.9cm,height=1.8cm]{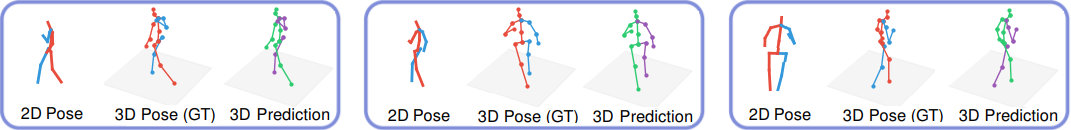} 
 \vspace*{-0.5cm}
\end{center}
   \caption{Visualization of 3D output of the estimation stage with some samples on the test set of Human3.6M \cite{6682899}. For each example, from left to right are 2D poses, 3D ground truths and our 3D predictions, respectively.}
\label{fig:5}
\end{figure}

We evaluate the effectiveness of the proposed 3D pose estimation network using the standard protocol of the Human3.6M dataset \cite{6682899,pavlakos2017coarse,martinez_2017_3dbaseline,VNect_SIGGRAPH2017}. Five subjects S1, S5, S6, S7, S8 are used for training and the rest two subjects S9, S11 are used for evaluation. Experimental results are reported by the average error in millimeters between the ground truth and the corresponding predictions over all joints. Much to our surprise, our method outperforms the previous best result from the literature \cite{martinez_2017_3dbaseline} by 3.1mm, corresponding to an error reduction of 6.8\% even when combining the ground truth 2D locations with the 2D OpenPose detections. This result proves that our network design can learn to recover the 3D pose from the 2D joint locations with a remarkably low error rate, which to the best of our knowledge, has established a new state-of-the-art on 3D human pose estimation (see Table \ref{tab:1} and Figure \ref{fig:5}). \\[-0.55cm]
\begin{table}[ht]
\caption{\label{tab:1} Experimental results and comparison with previous state-of-the-art 3D pose estimation approaches on the Human3.6M dataset \cite{6682899}.}
\vspace*{-0.2cm}
\begin{center}
\begin{tabular}{p{1.5cm}p{0.25cm}p{0.25cm}p{0.25cm}p{0.25cm}p{0.25cm}p{0.25cm}p{0.25cm}p{0.25cm}p{0.25cm}p{0.25cm}p{0.25cm}p{0.25cm}p{0.25cm}p{0.25cm}p{0.25cm}p{0.25cm}}
\hline
\hline
{\tiny Method} &  {\tiny Direct.} & {\tiny Disc.} & {\tiny Eat} & {\tiny Greet} & {\tiny Phone} & {\tiny Photo} & {\tiny Pose} & {\tiny Purch.} & {\tiny Sit} & {\tiny SitD} & {\tiny Smoke} & {\tiny Wait} & {\tiny WalkD} & {\tiny Walk} & {\tiny WalkT} & {\tiny Avg} \\
\hline
 \cellcolor{gray!30} {\tiny Ionescu \etal \cite{6682899}$^{\dagger}$} & \cellcolor{gray!30} {\tiny 132.7} & \cellcolor{gray!30} {\tiny 183.6} & \cellcolor{gray!30} {\tiny 132.3} & \cellcolor{gray!30} {\tiny 164.4} & \cellcolor{gray!30} {\tiny 162.1} & \cellcolor{gray!30} {\tiny 205.9} & \cellcolor{gray!30} {\tiny 150.6} & \cellcolor{gray!30} {\tiny 171.3} & \cellcolor{gray!30} {\tiny 151.6} & \cellcolor{gray!30} {\tiny 243.0} & \cellcolor{gray!30} {\tiny 162.1} & \cellcolor{gray!30} {\tiny 170.7} & \cellcolor{gray!30} {\tiny 177.1} & \cellcolor{gray!30} {\tiny 96.6} & \cellcolor{gray!30} {\tiny 127.9} & \cellcolor{gray!30} {\tiny 162.1} \\
 {\tiny Du \etal \cite{10.1007/978-3-319-46493-0_2}$^{\star}$} & {\tiny 85.1} & {\tiny 112.7} & {\tiny 104.9} & {\tiny 122.1} & {\tiny 139.1} & {\tiny 135.9} & {\tiny 105.9} & {\tiny 166.2} & {\tiny 117.5} & {\tiny 226.9} & {\tiny 120.0} & {\tiny 117.7} & {\tiny 137.4} & {\tiny 99.3} & {\tiny 106.5} & {\tiny 126.5} \\
 \cellcolor{gray!30} {\tiny Tekin \etal \cite{tekin2016direct} } & \cellcolor{gray!30} {\tiny 102.4} & \cellcolor{gray!30} {\tiny 147.2} & \cellcolor{gray!30} {\tiny88.8} & \cellcolor{gray!30} {\tiny 125.3} & \cellcolor{gray!30} {\tiny 118.0} & \cellcolor{gray!30} {\tiny 182.7} & \cellcolor{gray!30} {\tiny112.4} & \cellcolor{gray!30} {\tiny129.2} & \cellcolor{gray!30} {\tiny138.9} & \cellcolor{gray!30} {\tiny 224.9} & \cellcolor{gray!30} {\tiny 118.4} & \cellcolor{gray!30} {\tiny138.8} & \cellcolor{gray!30} {\tiny 126.3} & \cellcolor{gray!30} {\tiny 55.1} & \cellcolor{gray!30} {\tiny 65.8} & \cellcolor{gray!30} {\tiny 125.0} \\
 {\tiny Park \etal \cite{10.1007/978-3-319-49409-8_15}$^{\star}$} & {\tiny 100.3} & {\tiny 116.2} & {\tiny 90.0} & {\tiny 116.5} & {\tiny 115.3} & {\tiny 149.5} & {\tiny 117.6} & {\tiny 106.9} & {\tiny 137.2} & {\tiny 190.8} & {\tiny 105.8} &{\tiny 125.1} & {\tiny 131.9} & {\tiny 62.6} & {\tiny 96.2} & {\tiny 117.3}\\
 \cellcolor{gray!30} {\tiny Zhou \etal \cite{780906}$^{\star}$} &  \cellcolor{gray!30} {\tiny 87.4} & \cellcolor{gray!30} {\tiny 109.3} & \cellcolor{gray!30} {\tiny 87.1} & \cellcolor{gray!30} {\tiny 103.2} & \cellcolor{gray!30} {\tiny 116.2} & \cellcolor{gray!30} {\tiny 143.3} & \cellcolor{gray!30} {\tiny 106.9} & \cellcolor{gray!30} {\tiny 99.8} & \cellcolor{gray!30} {\tiny 124.5} & \cellcolor{gray!30} {\tiny 199.2} & \cellcolor{gray!30} {\tiny 107.4} & \cellcolor{gray!30} {\tiny 118.1} & \cellcolor{gray!30} {\tiny 114.2} & \cellcolor{gray!30} {\tiny 79.4} & \cellcolor{gray!30} {\tiny 97.7} & \cellcolor{gray!30} {\tiny 113.0} \\
{\tiny Zhou \etal \cite{Zhou2016DeepKP}$^{\star}$} & {\tiny 91.8} & {\tiny 102.4} & {\tiny 96.7} & {\tiny 98.8} & {\tiny 113.4} & {\tiny 125.2} & {\tiny 90.0} & {\tiny 93.8} & {\tiny 132.2} & {\tiny 159.0} & {\tiny 107.0} & {\tiny 94.4} & {\tiny 126.0} & {\tiny 79.0} & {\tiny 99.0} & {\tiny 107.3}\\
\cellcolor{gray!30} {\tiny Pavlakos \etal \cite{pavlakos2017coarse}} & \cellcolor{gray!30} {\tiny 67.4} & \cellcolor{gray!30} {\tiny 71.9} & \cellcolor{gray!30} {\tiny 66.7} & \cellcolor{gray!30} {\tiny 69.1} &  \cellcolor{gray!30} {\tiny 72.0} &  \cellcolor{gray!30} {\tiny 77.0} &  \cellcolor{gray!30} {\tiny 65.0} &  \cellcolor{gray!30} {\tiny 68.3} &  \cellcolor{gray!30} {\tiny 83.7} &  \cellcolor{gray!30} {\tiny 96.5} & \cellcolor{gray!30} {\tiny 71.7} & \cellcolor{gray!30} {\tiny 65.8} &  \cellcolor{gray!30} {\tiny 74.9} & \cellcolor{gray!30} {\tiny 59.1} & \cellcolor{gray!30} {\tiny 63.2} & \cellcolor{gray!30} {\tiny 71.9} \\
{\tiny Mehta \etal \cite{8374605}$^{\star}$} & {\tiny 67.4} & {\tiny 71.9} & {\tiny 66.7} & {\tiny 69.1} & {\tiny 71.9} & {\tiny 65.0} & {\tiny 68.3} & {\tiny 83.7} & {\tiny 120.0} & {\tiny 66.0} & {\tiny 79.8} & {\tiny 63.9} & {\tiny 48.9} & {\tiny 76.8} & {\tiny53.7} & {\tiny 68.6} \\
\cellcolor{gray!30} {\tiny Martinez \etal \cite{martinez_2017_3dbaseline}$^{\star}$} & \cellcolor{gray!30} {\tiny 51.8} & \cellcolor{gray!30} {\tiny 56.2} & \cellcolor{gray!30} {\tiny 58.1} & \cellcolor{gray!30} {\tiny 59.0} & \cellcolor{gray!30} {\tiny 69.5} & \cellcolor{gray!30} {\tiny 55.2} & \cellcolor{gray!30} {\tiny 58.1} & \cellcolor{gray!30} {\tiny 74.0} & \cellcolor{gray!30} {\tiny 94.6} & \cellcolor{gray!30} {\tiny 62.3} & \cellcolor{gray!30} {\tiny 78.4} & \cellcolor{gray!30} {\tiny 59.1} & \cellcolor{gray!30} {\tiny 49.5} & \cellcolor{gray!30} {\tiny 65.1} & \cellcolor{gray!30} {\tiny 52.4} & \cellcolor{gray!30} {\tiny 62.9} \\
{\tiny Liang \etal \cite{LIANG20181}} & {\tiny 52.8} & {\tiny 54.2} & {\tiny 54.3} & {\tiny 61.8} & {\tiny 53.1} & {\tiny 53.6} & {\tiny 71.7} & {\tiny 86.7} & {\tiny 61.5} & {\tiny 53.4} & {\tiny 67.2} & {\tiny 54.8} & {\tiny 53.4} & {\tiny 47.1} & {\tiny 61.6} & {\tiny 59.1} \\
\cellcolor{gray!30} {\tiny Luvizon \etal \cite{luvizon20182d}} & \cellcolor{gray!30} {\tiny 49.2} & \cellcolor{gray!30} {\tiny 51.6} & \cellcolor{gray!30} {\tiny 47.6} & \cellcolor{gray!30} {\tiny 50.5} & \cellcolor{gray!30} {\tiny 51.8} & \cellcolor{gray!30} {\tiny 48.5} & \cellcolor{gray!30} {\tiny 51.7} & \cellcolor{gray!30} {\tiny 61.5} & \cellcolor{gray!30} {\tiny 70.9} & \cellcolor{gray!30} {\tiny 53.7} & \cellcolor{gray!30} {\tiny 60.3} & \cellcolor{gray!30} {\tiny 48.9} & \cellcolor{gray!30} {\tiny 44.4} & \cellcolor{gray!30} {\tiny 57.9} & \cellcolor{gray!30} {\tiny 48.9} & \cellcolor{gray!30} {\tiny 53.2} \\
 {\tiny Martinez \etal \cite{martinez_2017_3dbaseline}$^{\dagger}$} &  {\tiny 37.7} &  {\tiny 44.4} &  {\tiny 40.3} &  {\tiny 42.1} &  {\tiny 48.2} &  {\tiny 54.9} &  {\tiny 44.4} &  {\tiny 42.1} &  {\tiny 54.6} &  {\tiny 58.0} &  {\tiny 45.1} &  {\tiny 46.4} &  {\tiny 47.6} &  {\tiny 36.4} &  {\tiny 40.4} &  {\tiny 45.5} \\
\hline
\cellcolor{gray!30} {\tiny \textbf{Ours}$^{\dagger,\star}$} & \cellcolor{gray!30} {\tiny \textbf{36.6}} & \cellcolor{gray!30} {\tiny \textbf{43.2}} & \cellcolor{gray!30} {\tiny \textbf{38.1}} & \cellcolor{gray!30} {\tiny \textbf{40.8}} & \cellcolor{gray!30} {\tiny \textbf{44.4}} & \cellcolor{gray!30} {\tiny \textbf{51.8}} & \cellcolor{gray!30} {\tiny \textbf{43.7}} & \cellcolor{gray!30} {\tiny \textbf{38.4}} & \cellcolor{gray!30} {\tiny \textbf{50.8}} & \cellcolor{gray!30} {\tiny \textbf{52.0}} & \cellcolor{gray!30} {\tiny \textbf{42.1}} & \cellcolor{gray!30} {\tiny \textbf{42.2}} & \cellcolor{gray!30} {\tiny \textbf{44.0}} & \cellcolor{gray!30} {\tiny \textbf{32.3}} & \cellcolor{gray!30} {\tiny \textbf{35.9}} & \cellcolor{gray!30} {\tiny \textbf{42.4}}\\
\hline
\hline \\[-0.75cm]                
\end{tabular}
\end{center}
\scriptsize{\hspace*{0.25cm} The symbol $^{\star}$ denotes that a 2D detector was used and the symbol $^{\dagger}$ denotes the ground truth 2D joint locations were used.} \\[-0.3cm]
\end{table} 
\subsubsection{Evaluation on action recognition \\[-0.1cm]}
\begin{figure}[ht]
\begin{center}
\includegraphics[width=5.5cm,height=4.5cm]{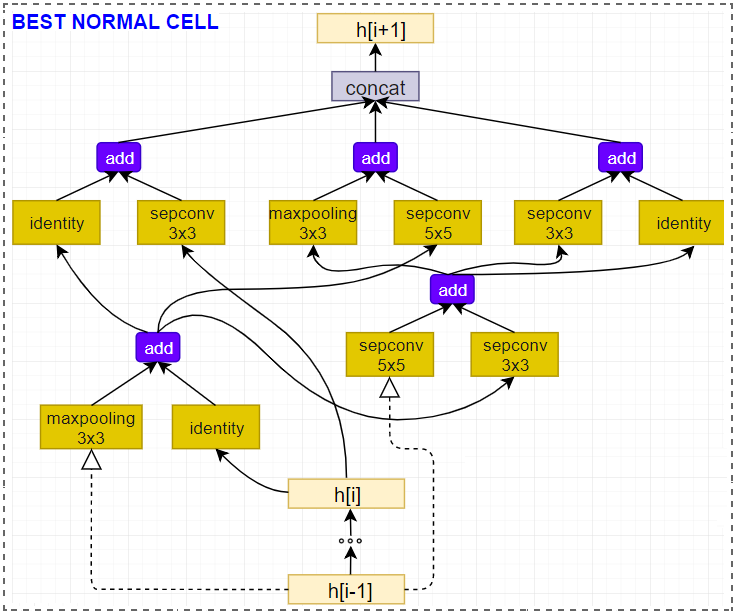}\includegraphics[width=5.5cm,height=4.5cm]{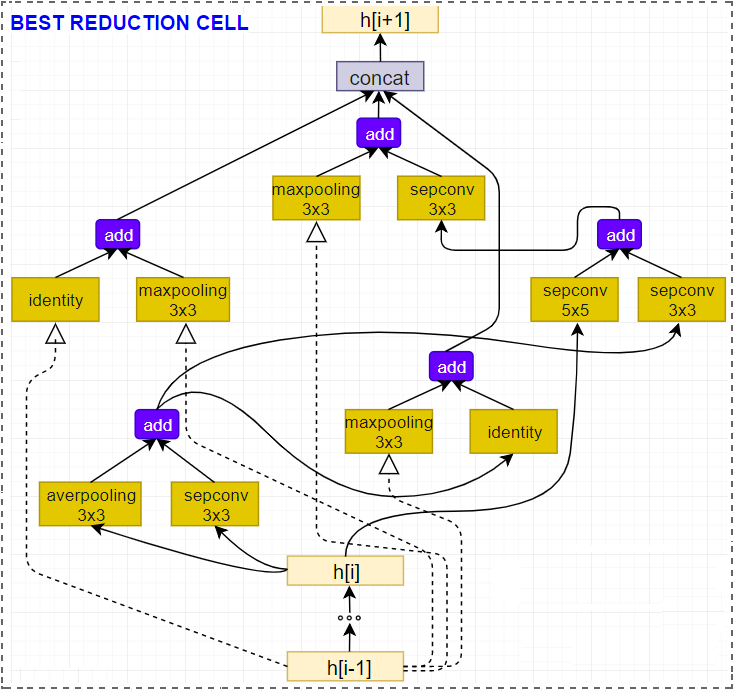} \hspace{0.1cm} \includegraphics[width=1.5cm,height=4.5cm]{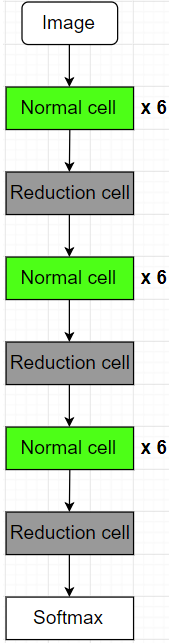}\\
\scriptsize{ \hspace*{2cm} (a) \hspace{5.2cm} (b) \hspace{3cm} (c)}
\vspace*{-0.5cm}
\end{center}
   \caption{Diagram of the top performing \textit{normal cell} (a) and \textit{reduction cell} (b) discovered by ENAS \cite{pmlr-v80-pham18a} on AS1 subset \cite{li2010action}. They were then used to construct the final network architecture (c). We recommend the interested readers to \cite{pmlr-v80-pham18a} to better understand this procedure.} 
\label{fig:6}
\end{figure}

Table~\ref{table2} reports the experimental results and comparisons with state-of-the-art methods on the MSR Action3D dataset \cite{li2010action}. The ENAS algorithm \cite{pmlr-v80-pham18a} is able to explore a diversity of network architectures and the best design is identified based on its validation score. Thus, the final architecture achieved a total average accuracy of 97.98\% over three subset AS1, AS2 and AS3. This result outperforms many previous studies \cite{li2010action,Chen2013RealtimeHA,Vemulapalli2014HumanAR,Du2015HierarchicalRN,Liu2016SpatioTemporalLW,Wang2016GraphBS,WengSpatioTemporalNN,Xu2015SpatioTemporalPM,lee2017ensemble}, and among them, many are depth sensor-based approaches. Figure \ref{fig:6} provides a schematic diagram of the best cells and optimal architecture found by ENAS on the AS1 subset \cite{li2010action}. For the SBU Kinect Interaction dataset \cite{yun2012two}, the best model achieved an accuracy of 96.30\%, as shown in Table~\ref{table3}. Our reported results indicated an important observation that by using only the 3D predicted poses, we are able to outperform previous works reported in \cite{Song2017AnES,Liu2016SpatioTemporalLW,8322204,ke2017new,DBLP:conf/bmvc/TasK18,Wang2017ModelingTD,liu2018skeleton} and reach state-of-the-art results provided in \cite{zhang2019view,s19081932}, which deploy accurate skeletal data provided by Kinect v2 sensor.
\begin{table}
\parbox{.45\linewidth}{
\centering
\caption{Test accuracies (\%) on the MSR Action3D dataset \cite{li2010action}.}
\vspace*{-0.3cm}
\begin{tabular}{ccccc}
\label{table2}\\
\hline
{\scriptsize \hspace*{0.01cm} \textbf{Method}} & {\scriptsize \hspace*{0.01cm}  \textbf{AS1}} & {\scriptsize \hspace*{0.01cm} \textbf{AS2}} & {\scriptsize \hspace*{0.01cm} \textbf{AS3}} &  {\scriptsize \hspace*{0.01cm} \textbf{Aver.}}  \\
 \hline
 {\scriptsize \hspace*{0.01cm} Li \etal \cite{li2010action}} & {\scriptsize \hspace*{0.01cm} 72.90} & {\scriptsize \hspace*{0.01cm} 71.90}  & {\scriptsize \hspace*{0.01cm} 71.90} &  {\scriptsize \hspace*{0.01cm} 74.70} \\
{\scriptsize \cellcolor{gray!30}  Chen \etal \cite{Chen2013RealtimeHA}} & {\scriptsize \cellcolor{gray!30} 96.20} & {\scriptsize \cellcolor{gray!30} 83.20}  & {\scriptsize \cellcolor{gray!30} 92.00} &  {\scriptsize \cellcolor{gray!30} 90.47} \\
{\scriptsize \hspace*{0.01cm}  Vemulapalli \etal \cite{Vemulapalli2014HumanAR}} & {\scriptsize \hspace*{0.01cm}  95.29} & {\scriptsize \hspace*{0.01cm}  83.87}  & {\scriptsize  \hspace*{0.01cm} 98.22} &  {\scriptsize \hspace*{0.01cm} 92.46} \\
{\scriptsize \cellcolor{gray!30} Du \etal \cite{Du2015HierarchicalRN}} & {\scriptsize \cellcolor{gray!30}    99.33} & {\scriptsize \cellcolor{gray!30}  94.64}  & {\scriptsize \cellcolor{gray!30}   95.50} &  {\scriptsize \cellcolor{gray!30}  94.49} \\
{\scriptsize  \hspace*{0.01cm} Liu \etal \cite{Liu2016SpatioTemporalLW}} & {\scriptsize \hspace*{0.01cm} N/A} & {\scriptsize  \hspace*{0.01cm} N/A}  & {\scriptsize  \hspace*{0.01cm} N/A} &  {\scriptsize  \hspace*{0.01cm} 94.80}
 \\
{\scriptsize \cellcolor{gray!30} Wang \etal \cite{Wang2016GraphBS}} & {\scriptsize \cellcolor{gray!30}   93.60} & {\scriptsize  \cellcolor{gray!30}  95.50}  & {\scriptsize  \cellcolor{gray!30}   95.10} &  {\scriptsize \cellcolor{gray!30}   94.80}
 \\
{\scriptsize  \hspace*{0.01cm} Wang \etal \cite{WengSpatioTemporalNN}} & {\scriptsize  \hspace*{0.01cm} 91.50} & {\scriptsize   \hspace*{0.01cm} 95.60}  & {\scriptsize   \hspace*{0.01cm} 97.30} &  {\scriptsize  \hspace*{0.01cm} 94.80} \\
{\scriptsize \cellcolor{gray!30} Xu \etal \cite{Xu2015SpatioTemporalPM}} & {\scriptsize\cellcolor{gray!30} 99.10} & {\scriptsize \cellcolor{gray!30} 92.90}  & {\scriptsize  \cellcolor{gray!30} 96.40} &  {\scriptsize \cellcolor{gray!30} 96.10} \\
{\scriptsize \hspace*{0.01cm} Lee \etal \cite{lee2017ensemble}} & {\scriptsize  \hspace*{0.01cm} 95.24} & {\scriptsize  \hspace*{0.01cm} 96.43}  & {\scriptsize   \hspace*{0.01cm}  100.0} &  {\scriptsize  \hspace*{0.01cm} 97.22} \\
{\scriptsize \cellcolor{gray!30} \hspace*{0.01cm} Pham \etal \cite{s19081932}} & {\scriptsize \cellcolor{gray!30}  \hspace*{0.01cm} 98.83} & {\scriptsize  \cellcolor{gray!30} \hspace*{0.01cm} 99.06}  & {\scriptsize  \cellcolor{gray!30} \hspace*{0.01cm}  99.40} &  {\scriptsize  \cellcolor{gray!30} \hspace*{0.01cm} 99.10} \\
\hline
{\scriptsize  \hspace*{0.01cm} \textbf{Ours}} & {\scriptsize   \hspace*{0.01cm} \textbf{97.87} } & {\scriptsize   \hspace*{0.01cm} \textbf{96.81}}  & {\scriptsize   \hspace*{0.01cm} \textbf{99.27}} &  {\scriptsize  \hspace*{0.01cm} \textbf{97.98}} \\    
\hline
\hline
\end{tabular}}
\hfill
\parbox{.45\linewidth}{
\centering
\caption{Test accuracies (\%) on the SBU Kinect Interaction dataset \cite{yun2012two}.}
\vspace*{-0.3cm}
\begin{tabular}{cc}
\label{table3}\\
\hline
\hspace*{0.01cm} {\scriptsize \textbf{Method}} & {\scriptsize \hspace*{0.3cm} \textbf{Acc.}} \\
\hline
{\scriptsize \hspace*{0.01cm} Song \etal \cite{Song2017AnES}} & {\scriptsize \hspace*{0.32cm} 91.51} \\
{\scriptsize \cellcolor{gray!30} Liu \etal \cite{Liu2016SpatioTemporalLW}} & {\scriptsize \cellcolor{gray!30}  \hspace*{0.25cm} 93.30}\\
{\scriptsize  \hspace*{0.01cm}  Weng \etal \cite{8322204}} & {\scriptsize   \hspace*{0.32cm} 93.30} \\
{\scriptsize   \cellcolor{gray!30} Ke \etal \cite{ke2017new}} & {\scriptsize \cellcolor{gray!30}  \hspace*{0.25cm} 93.57} \\
{\scriptsize   \hspace*{0.01cm} Tas \etal \cite{DBLP:conf/bmvc/TasK18}} & {\scriptsize  \hspace*{0.32cm} 94.36} \\
{\scriptsize \cellcolor{gray!30}   \hspace*{0.01cm} Wang \etal  \cite{Wang2017ModelingTD}} & {\scriptsize  \cellcolor{gray!30} \hspace*{0.32cm} 94.80} \\
{\scriptsize   \hspace*{0.01cm}  Liu \etal \cite{liu2018skeleton}} & {\scriptsize  \hspace*{0.32cm} 94.90}  \\
{\scriptsize \cellcolor{gray!30}   \hspace*{0.01cm} Zang \etal  \cite{zhang2019view} (using VA-RNN)} & {\scriptsize  \cellcolor{gray!30} \hspace*{0.32cm} 95.70} \\
{\scriptsize \hspace*{0.01cm} Zhang \etal  \cite{zhang2019view} (using VA-CNN)} & {\scriptsize   \hspace*{0.32cm} 97.50} \\
{\scriptsize \cellcolor{gray!30} \hspace*{0.01cm} Pham \etal  \cite{s19081932}} & {\scriptsize \cellcolor{gray!30}  \hspace*{0.32cm} 97.86} \\
\hline
{\scriptsize    \textbf{Ours} } & {\scriptsize   \hspace*{0.4cm} \textbf{96.30}}  \\
\hline
\hline
\end{tabular}}
\end{table}
\subsection{Computational efficiency evaluation  \\[-0.1cm]} \label{sect:4.4}
On a single GeForce GTX 1080Ti GPU  with 11GB memory, the runtime of OpenPose \cite{cao2017realtime} is less than 0.1\textit{s} per frame on a image size of 800 $\times$ 450 pixels. On the Human3.6M dataset \cite{6682899}, the 3D pose estimation stage takes around 15\textit{ms} to complete a pass (forward + backward) through each stream with a mini-batches of size 128. Each epoch was done within 3 minutes. For the action recognition stage, our implementation of ENAS algorithm takes about 2 hours to find the final architecture ($\sim$2.3M parameters) on each subset of MSR Action3D dataset \cite{li2010action}, whilst it takes around 3 hours on the SBU Kinect Interaction dataset \cite{yun2012two} to discover the best architecture ($\sim$3M parameters). With small architecture sizes, the discovered networks require low computing time for the inference stage, making our approach more practical for large-scale problems and real-time applications. 
\section{Conclusions \\[-0.2cm]} \label{sect:5}
In this paper, we presented a unified deep learning framework for joint 3D human pose estimation and action recognition from RGB video sequences. The proposed method first runs a state-of-the-art 2D pose detector to estimate 2D locations of body joints. A deep neural network is then designed and trained to learn a direct 2D-to-3D mapping and predict human poses in 3D space. Experimental results demonstrated that the 3D human poses can be effectively estimated by a simple network design and training methodology over 2D keypoints. We also introduced a novel action recognition approach based on a compact image-based representation and automated machine learning, in which an advanced neural architecture search algorithm is exploited to discover the best performing architecture for each recognition task. Our experiments on public and challenging action recognition datasets indicated that the proposed framework is able to reach state-of-the-art performance, whilst requiring less computation budget for training and inference. Despite that, our method naturally depends on the quality of the output of the 2D detectors. Hence, a limitation is that it cannot recover 3D poses from 2D failed output. To tackle this problem, we are currently expanding this study by adding more visual evidence to the network in order to further gains in performance. The preliminary results are encouraging. 
\bibliography{egbib}
\end{document}